\documentclass[letterpaper, 10 pt, conference]{ieeeconf}  
\IEEEoverridecommandlockouts                                      
\usepackage{amsmath}
\usepackage{amssymb}
\usepackage{mathtools}
\usepackage{graphicx}
\usepackage{float}
\usepackage{array}
\usepackage{tikz}
\usepackage{listings}
\usepackage{hyperref}
\usepackage{graphicx}
\usepackage{cite}
\usepackage{dsfont}
\usepackage{mathtools,etoolbox}
\usepackage[none]{hyphenat}
\usepackage[utf8]{inputenc}
\usepackage[english]{babel}

\usepackage{amsthm}
\usepackage{xfrac}
\usepackage{nicefrac}
\usepackage{booktabs}
\usepackage[switch, pagewise]{lineno}
\usepackage{nicefrac}
\usepackage{flushend}

\hypersetup{colorlinks=true,urlcolor=blue,citecolor=red}

\usepackage{colortbl}
\usepackage{multirow}
\usepackage{color, colortbl}
\definecolor{LightCyan}{rgb}{0.88,1,1}

\makeatletter

\usepackage{algpseudocode}
\usepackage{algorithm}

\long\def\invis#1{}

\title{\LARGE \bf
\textit{OysterNet}: Enhanced Oyster Detection Using Simulation
\thanks{This work was supported by USDA NIFA sustainable agriculture system program under award
number 20206801231805.}}
\author{Xiaomin Lin$^{1}$,  Nitin J. Sanket$^{2}$, Nare Karapetyan$^{1}$, Yiannis Aloimonos$^{1}$ 
\thanks{$^{1}$Perception and Robotics Group, University of Maryland Institute for Advanced Computer Studies, University of Maryland, College Park, MD 20742, USA. Emails: \texttt{\{xlin01, knare, jyaloimo\}@umd.edu}.
        }%
\thanks{$^{2}$Robotics Engineering, Worcester Polytechnic Institute, MA 01609, USA. Email: \texttt{nsanket@wpi.edu}.
        }%
}

\begin{document}

\maketitle
\thispagestyle{empty}
\pagestyle{empty}


\begin{abstract}
Oysters play a pivotal role in the bay living ecosystem and are considered the living filters for the ocean. In recent years, oyster reefs have undergone major devastation caused by commercial over-harvesting, requiring preservation to maintain ecological balance. The foundation of this preservation is to estimate the oyster density which requires accurate oyster detection.
However, systems for accurate oyster detection require large datasets obtaining which is an expensive and labor-intensive task in underwater environments. 

To this end, we present a novel method to mathematically model oysters and render images of oysters in simulation to boost the detection performance with minimal real data. 
Utilizing our synthetic data along with real data for oyster detection, we obtain up to 35.1\% boost in performance as compared to using only real data with our \textit{OysterNet} network. We also improve the state-of-the-art by 12.7\%.
This shows that using underlying geometrical properties of objects can help to enhance recognition task accuracy on limited datasets successfully and we hope more researchers adopt such a strategy for hard-to-obtain datasets.

\end{abstract}

\section{INTRODUCTION}
\label{section:introduction}



Oyster reefs are filter feeders and they provide crucial benefits for the benthic marine ecosystem(s) such as increasing
the richness for a variety of species, providing living habitat, food, and protection for numerous marine species. However, the standing stocks for the oysters near the Chesapeake Bay\cite{newell1988ecological} and North Sea\cite{pogoda2019current} have dropped significantly due to  over-fishing, global warming and the effect of diseases across the 19th century. To tackle this devastating ecological problem, massive efforts are being carried out to restore oyster habitats across the United States \cite{baggett2015guidelines, blomberg2018habitat, mcfarland2018restoring, theuerkauf2019integrating} and Europe \cite{pogoda2019current}.



One of the core challenges to advance, improve and adapt the restoration process is monitoring of the progress of oyster restorations effectively. Beck et al.~\cite{beck2011oyster} proposed to standardize monitoring
metrics, units, and performance criteria for the evaluation of the oyster reefs. A set of environmental variables including water salinity, temperature, and dissolved oxygen are being monitored to determine the well-being of oyster habits. For the oyster reefs, universal parameters such as ``reef areal dimensions, reef height, oyster density, and oyster size-frequency distribution'' are monitored and reported in the literature \cite{beck2011oyster,pogoda2019current, blomberg2018habitat, mcfarland2018restoring, theuerkauf2019integrating}. \invis{These metrics and variables are also monitored and reported in previous works\cite{pogoda2019current}, Blomberg\cite{blomberg2018habitat},McFarland \cite{ mcfarland2018restoring}, and Theuerkauf\cite{theuerkauf2019integrating}.}
\begin{figure}
 \centering
{\includegraphics[width=1.0\linewidth]{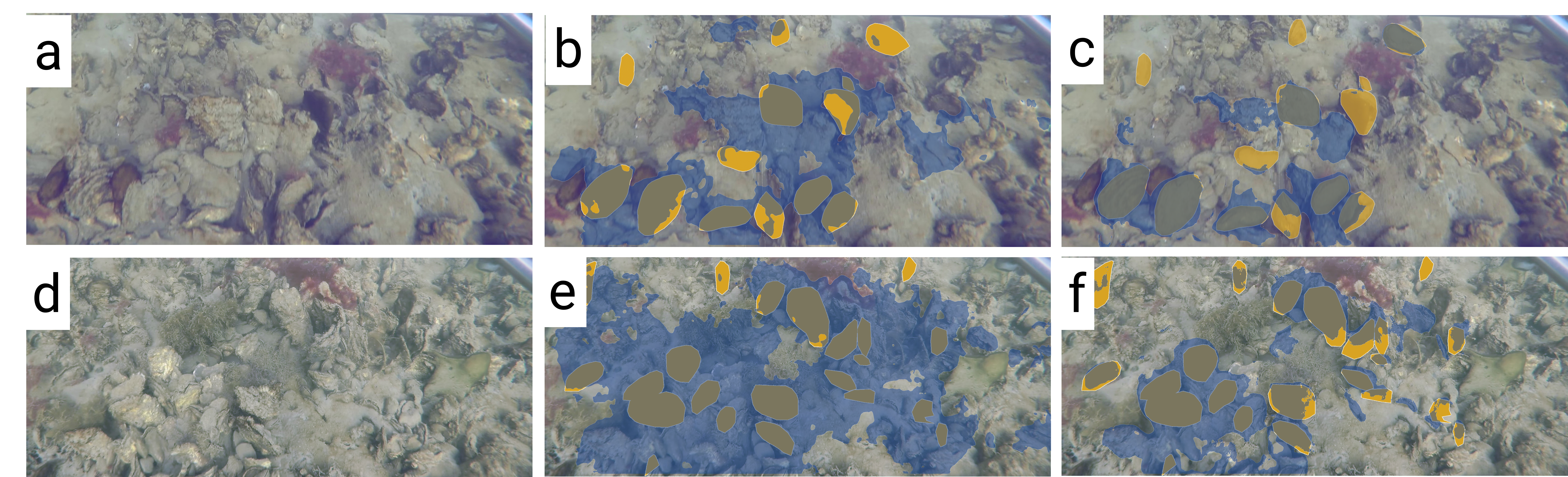}}
\caption{Each row left to right: Input image, output of the network when trained using only real data, output of the network (which we call \textit{OysterNet}) when trained using real data augmented with our synthetic data. Yellow represents the oyster segmentation ground truth and the blue is the predicted segmentation result. Notice how the number of false positives and false negatives drop significantly when the training data is augmented with our synthetic data, \textit{All the images in this paper are best seen in color on a computer screen at 200\% zoom.}}

\label{fig:detection_result}
\end{figure}
However, these metrics rely on recognizing and counting oysters, which is currently largely done by expert manual labor. Such a process is slow, time-consuming, and has poor scalability. 
Using such a manual approach, the oyster reefs can only be monitored within a restricted area with few samples, e.g., 100 oysters per sample site \cite{mcfarland2018restoring}. Furthermore, the material for the oyster's surface is similar to the seabed sediment (See Fig.~\ref{fig:detection_result}\textcolor{red}{a}, Fig.~\ref{fig:detection_result}\textcolor{red}{d},). 
And it is significantly different from the washed oysters in Fig. \ref{fig:Real-2-Sim}\textcolor{red}{a}, which makes it very challenging to train new people or algorithms to perform oyster counting. 

To streamline the process of oyster mapping, the goal is to utilize the advancements in robotics and artificial intelligence that can enable us to gather images from underwater Remotely Operated Vehicles (ROVs) and then automate the oyster detection and density calculation. The central part of this process is to build an oyster detection system. In this work, we present a mathematical model to generate oyster models and further use Generative Adversarial Networks to enable sim-to-real transfer. To the best of our knowledge, this is the first attempt to geometrically model oysters. The contributions of this paper are as follows:


\begin{itemize}
\item We propose a novel mathematical model for the 3D shape of oysters.
\item We simplify the geometric model of an oyster for the projection on the image plane which is used to generate photorealistic synthetic oyster images. These images are used to train a deep segmentation network \textit{OysterNet} for oysters that achieves the new state-of-the-art.
\item We open-source our oyster generation model and dataset associated with this work to accelerate further research.
\end{itemize}

\begin{figure*}[ht!]
\includegraphics[width=\textwidth]{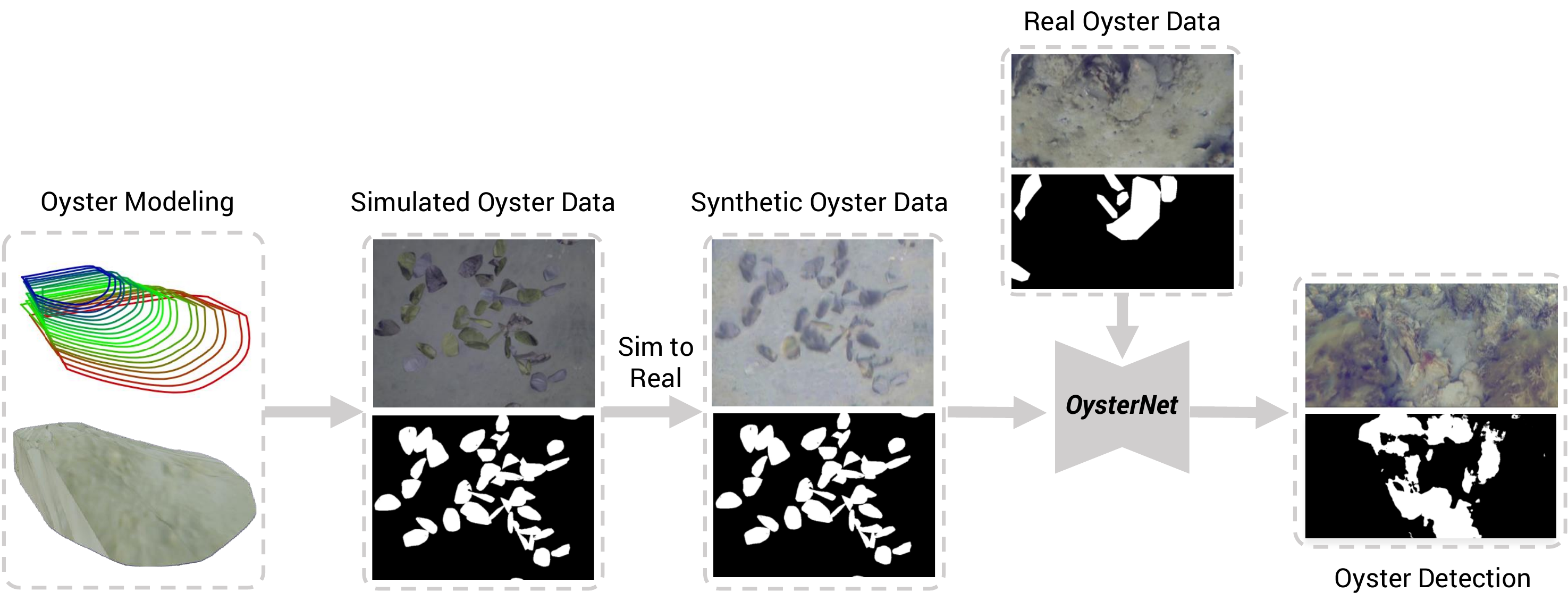}
\centering
\vspace{-5mm}
\caption{An overview of our approach: The proposed geometric model is used to generate synthetic images which are further fed into a Generative Adversarial Network to enable sim-to-real transfer (domain adaptation) by generating photorealistic oyster images. We then combine the synthetic data with real data to train a \textit{OysterNet} for oyster detection.}
\vspace{-5mm}
\label{fig:overview}
\end{figure*}
The rest of this paper is organized as follows: First, we place this work in the context of previous works in Sec.~\ref{section:related_work}. Then, we describe the proposed geometric model of the oyster which is used to create realistic images in Sec.~\ref{section:problem_formulation}.
We then present extensive quantitative and qualitative evaluations of our approach in Sec.~\ref{section:Experiments_and_results}. Finally, we conclude the paper in Sec.~\ref{section:Conclusions} with parting thoughts on future work.

\section{RELATED WORK}
\label{section:related_work}

Automation and robotics are becoming an integral part of many applications, particularly in the marine domain for environmental monitoring\cite{manjanna2016efficient,hansen2018autonomous, karapetyan2021meander}. The marine domain poses many additional challenges on top of the classical ones faced by robotics. Some of the commonly encountered problems are image visibility distortions caused by the water or sediment, and the difficulty of acquiring data for developing recognition or planning methods that are driven by the information. This is especially true for oyster monitoring. 

Systematic monitoring of underwater ecosystem requires reliable autonomous navigation. However, in underwater environments, navigation is challenging not only due to the dynamics of the systems but also due to a lack of adequate spatial awareness. To overcome these challenges vision-only based methods have been developed in the literature, that use human-labeled data to learn navigation commands for surveying coral reefs~\cite{manderson2018vision} and shipwrecks~\cite{karapetyan2021human}. \invis{On the other hand, perception-based navigation requires an abundance of data, which is hard to acquire for the underwater domain. To overcome Manderson et al.~\cite{manderson2018vision} developed a method for performing monitoring over a coral reef that uses human-labeled data to learn navigation commands. Motivated by the latter Karapetyan et al.~\cite{karapetyan2021human} proposed an underwater coverage method using human labeled navigation strategy for applications toward mapping shipwreck structures.} It is important to point out that these methods heavily depend on the quality of data and are hence require extensive data collection for a high degree of robustness.\invis{Once visual landmarks are identifiable it is possible to use a classical optimization-based planning framework to perform perception-aware navigation~\cite{xanthidis2021aquavis}.}

Sadrfaridpour et al.\cite{sadrfaridpour2021detecting} recently collected an underwater dataset and used Mask-R-CNN\cite{he2017mask} for oyster detection. However, this dataset is relatively small and is not collected in the ocean/sea bed but rather on an oyster farm. The oysters are stacked and are very dense in the dataset which lead to poor detection results for oyster reefs in the real world. There is a lack of variety of oysters and environment in Sadrfaridpour's dataset for a more robust detection result when deployed in the wild. 

Instead of collecting large datasets for detection tasks, we follow the conceptual approach proposed by 
Sanket et al.~\cite{sanket2021evpropnet}  which is to geometrically model the object under consideration to generate an enormous amounts of data synthetically. In particular, we model the 3D shape of oysters to create an underwater dataset for oyster detection.

\invis{In addition to the learning-based methods mentioned above,}

Alternatively, Generative Adversarial Networks (GANs) has been also widely used to generate underwater datasets. Joshi et al.~\cite{joshi2020deepurl} utilized a GAN to create images for pose estimation of an autonomous underwater vehicle (AUV). However, this method also required a large number of ground truth data samples for generating all possible pose images. To tackle the visibility distortions caused by the water or sediment, Li et al.~\cite{li2017watergan} proposed a system that restores underwater images into in-air images. Moreover, Wang~\cite{wang2019uwgan} also proposed an approach for real-world underwater color restoration and dehazing by utilizing GANs.\invis{ Rather than using a GAN, Akkaynak et al.~\cite{akkaynak2019sea} proposed an accurate physical model for removing the water effect from underwater images.}

With the ultimate goal of developing an autonomous information-driven oyster monitoring system, we acknowledge the need of good oyster prediction methods. Moreover due to the lack of large samples of oyster detests and the lack of literature to address this problem we are seeking an alternative path for generating a large oyster dataset by looking into the mathematical model of oysters. To the best of our knowledge, we are the first to propose a geometric model of oysters and use that model to generate synthetic data which is described next.
\section{Synthetic Oyster Generation}
\begin{figure*}[ht!]
\includegraphics[width=\textwidth]{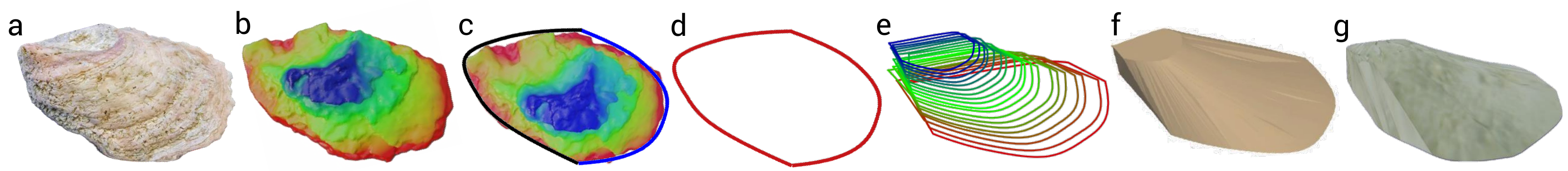}
\centering
\vspace{-5mm}
\caption{Steps in the geometric modelling of an oyster:  (a) Sample image of a real oyster shell, (b) 3D scan of a real oyster, (c) Splines fit to the oyster's bottom layer to model it (each color represents a single spline), (d) Simplified model of one stratified layer of the oyster (single layer of spline curve $S(t)$), (e) all layers of $S^\alpha(t)$, (f) generated 3D model of oyster, (g) final generated 3D model of oyster with added real oyster texture.}
\label{fig:Real-2-Sim}
\end{figure*}
\label{section:problem_formulation}
In this work, we propose a novel method for modeling oysters which are then used to generate a large dataset of oyster reef images. In this section, we will talk about our proposed method (which is summarised in Fig.~\ref{fig:overview}). First, we will describe the process of modeling oysters. Next, we will talk about how this model can be used to generate photorealistic images of oyster reefs using domain adaptation. Last but not least, we train a semantic segmentation model using the \textit{OysterNet} for detecting the oysters in the real world. We present each sub-part in the following sub-sections.



\subsection{Geometric Modelling Of Oysters}

In order to model the geometry of an oyster, we first 3D scanned ten washed oysters (sample image example shown in \ref{fig:Real-2-Sim}\textcolor{red}{a} and scanned model shown in \ref{fig:Real-2-Sim}\textcolor{red}{b}) which were used to build a mathematical model. Furthermore, the parameters in the proposed mathematical model can be adjusted to generate models of various oysters. We will describe the mathematical model next.



It is important to note that the geometrical shape of an oyster is extremely complex, with each one having a different shape and thickness. Oysters grow following a general oyster shape but expand somewhat randomly along their margins. 
In the first step, we will model our oysters in 3D. Each oyster can be approximated as a series of mathematical functions in a stratified manner (akin to the way a 3D printer prints). We start with each `horizontal' layer (slice of the cross-section of the oyster) by looking at the top view of our scanner oysters. We call this the perimeter of the oyster. We noticed that the perimeter can be easily modeled using two cubic B-splines. Let the $n+1$ control points for the splines be $\{c_0, ... , c_n\}$ and $m + 1$ knot vectors be ${\{t_0, ..., t_m\}}$, then the spline curve $S(t)$ of degree $k$ is given by

\begin{equation}
	S(t) = \sum_{i=0}^{n} c_i B_{i,k}(t) = 1,
	\label{eq:S_t}
\end{equation}
where, $B_{i,k}(t)$ denotes the basic function of degree $k$ and is computed recursively as

\begin{equation}
	B_{i,0}(t) =
\begin{cases}
1         & \text{if } t_{i+1}\geq t\geq t_{i}\\
0,        & \text{otherwise},
\end{cases}
\label{eq:B_t_0}
\end{equation}

\begin{equation}
	B_{i,k}(t) = \frac{t-t_i}{t_{i+k}-t_i}B_{i,k-1}(t)+ \frac{t_{i+k+1}-t}{t_{i+k+1}-t_{i+1}}B_{i+1,k-1}(t).
	\label{eq:B_t_k}
\end{equation}
In our case here, $m = n + k + 1$ and we selected $k=3$ for a cubic spline. Particularly, we utilize two cubic B-splines: one for the top half of the shell, and one for the bottom of the shell (See Figs. \ref{fig:Real-2-Sim}\textcolor{red}{c} and \ref{fig:Real-2-Sim}\textcolor{red}{d}). We will call this perimeter model (Eqs. \ref{eq:S_t}, \ref{eq:B_t_0} and \ref{eq:B_t_k}) as the 2D model since it models only a single layer of the oyster.

Now, we want to extend the 2D model to 3D in the stratified manner we described before. However, having a high resolution in depth is computationally prohibitive due to all the nooks and crannies (high-frequency edges) on the oyster shell (See Fig. \ref{fig:Real-2-Sim}\textcolor{red}{b}). In order to make this computation tractable, we simplify the shape of the oyster by assuming that it is smooth since the variation on the shell is much smaller than the distance to the oyster. It is important to highlight that the visual changes these high-frequency edges created are approximated by the visual textures we place on our generated oysters. The 3D model of the oyster (Fig. \ref{fig:Real-2-Sim}\textcolor{red}{e}) follows the same perimeter model from before but also adds changes to depth. Let $c^\alpha_0, ... , c^\alpha_n$ denote the control points for $\alpha$-th layer of the 3D oyster. And the knot vectors for the $\alpha$-th layer be${\{t^\alpha_0,...,t^\alpha_m\}}$. \invis{$\alpha$ is a number between 15 and 20.} We define the ${c^\alpha_n}$,$t^\alpha_m$ as follows
\begin{equation}
\vspace{-3mm}
{c^\alpha_n} ={c_n} + X_1 
\label{eq:c_an}
\end{equation} 

\begin{equation}
{t^\alpha_m} = t_m + X_2
\label{eq:t_am}
\end{equation}
where $X1$ and $X2$ are the Gaussian Noise used to model the high frequency edges of the oyster's perimeter. Formally, $X1 \sim \mathcal{N}(\mu_1,\,\sigma_1^{2})$ and 
$X2 \sim \mathcal{N}(\mu_2,\,\sigma_2^{2})$.

We can then substitute $c_n,t_m$ with ${c^\alpha_n},{t^\alpha_m}$ in Eqs. \ref{eq:S_t} to \ref{eq:B_t_k} to get the spline curve $S^\alpha(t)$ for every single layer. 

For each spline curve $S^\alpha(t)$, we can use $[x^\prime_\alpha$, $y^\prime_\alpha]^T$ to represent spline curve points as follows:
\begin{equation}
[x_\alpha, y_\alpha]^T =  [x^\prime_\alpha,  y^\prime_\alpha]^Tr^\alpha
\end{equation}
where $r$ is the in-growth rate as a fixed number.

Finally, the growth rate of the oyster is defined as follows:\invis{The upper layer of the perimeter would grow inward:}
\begin{equation}
z_\alpha = \alpha d,
\end{equation}
where $d$ is the depth at a fixed number here.

Now we can use all the  $\alpha$ layers with $[x_\alpha,y_\alpha,z_\alpha]^T$ points to generate 3D model (Fig.~\ref{fig:Real-2-Sim}\textcolor{red}{f}) by using pyvista~\cite{sullivan2019pyvista} data visualizer. By varying the parameters $\Theta=\{\sigma_1, \sigma_2, \mu_1, \mu_2, \alpha\}$ we obtain different oyster shapes (Fig. \ref{fig:ablation_study}). Further, we use image textures from real oysters collected in the Chesapeake Bay to be warped on the generated 3D model (See Fig. \ref{fig:Real-2-Sim}\textcolor{red}{g}).  

\begin{figure}[b!]
\includegraphics[width=\linewidth]{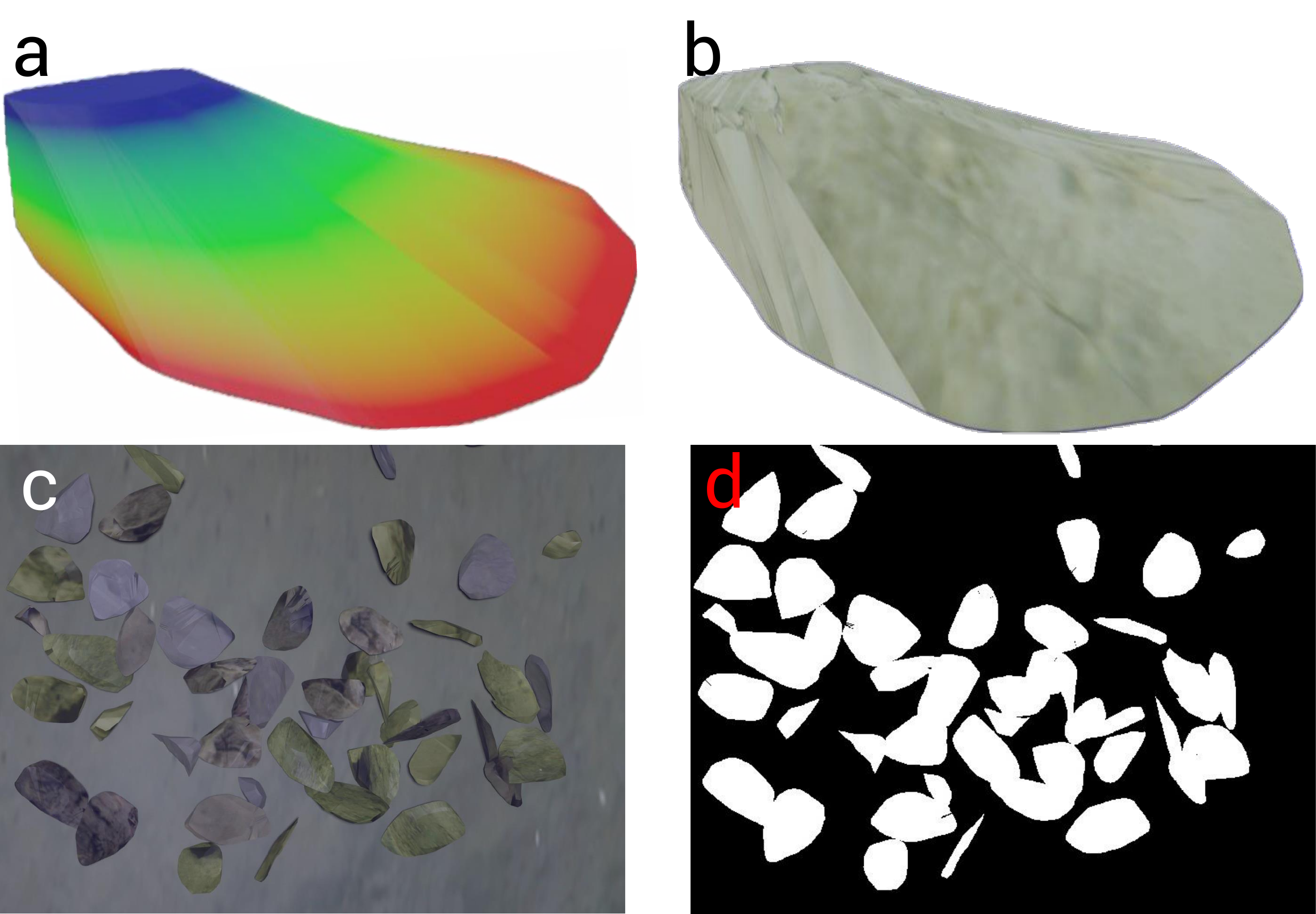}
\centering
\caption{(a) One of the synthetic models generated from Sec. \ref{section:problem_formulation}\textcolor{red}{-A}),  (b) Synthetic model with real oyster texture added, (c) image of an oyster farm with 50 synthetic oysters generated in Blender$^{\text{TM}}$, (d) masks for oysters in (c).}
\label{fig:simulation_images}
\end{figure}

In the next section, we will talk about how the actual images are rendered from the 3D models we just constructed.



\subsection{Synthetic Image Generation}

\subsubsection{Simulation Image Rendering} 
A 3D model is not sufficient for creating images for oyster segmentation. We utilized the Blender$^{\text{TM}}$ \cite{blender} game engine to simulate the oysters on a seabed. We rendered 13K synthetic oysters with different 3D models (we will just call it the synthetic model for simplicity) by varying parameters $\Theta$ as described in Sec. \ref{section:problem_formulation}. Then we used these synthetic 3D models for synthetic data generation. First, we employed 14 real oyster texture images and applied them randomly to all the synthetic models (Fig.~\ref{fig:simulation_images}\textcolor{red}{a}). A sample of the synthetic model with applied texture image can be seen in Fig.~\ref{fig:simulation_images}\textcolor{red}{b}. To generate an simulated oyster reef image, we placed the oysters with random poses onto a flat surface with the seabed textures as shown in  Fig~\ref{fig:simulation_images}\textcolor{red}{c}. Images of the oyster reef are then rendered in Blender$^{\text{TM}}$ along with segmentation masks (Fig.~\ref{fig:simulation_images}\textcolor{red}{d}).

However, Fig.~\ref{fig:simulation_images}\textcolor{red}{c} is not photorealistic and does not look like a real underwater image which will lead to a poor detection performance when deployed. We describe how we use a Generative Adversarial Network (GAN) to perform sim-to-real domain adaptation such that our images are photorealistic such that they can generalize to the real world after being trained in simulation.\\[-5pt]


\subsubsection{Domain Adaptation}
\begin{figure}[b!]
\includegraphics[width=\linewidth]{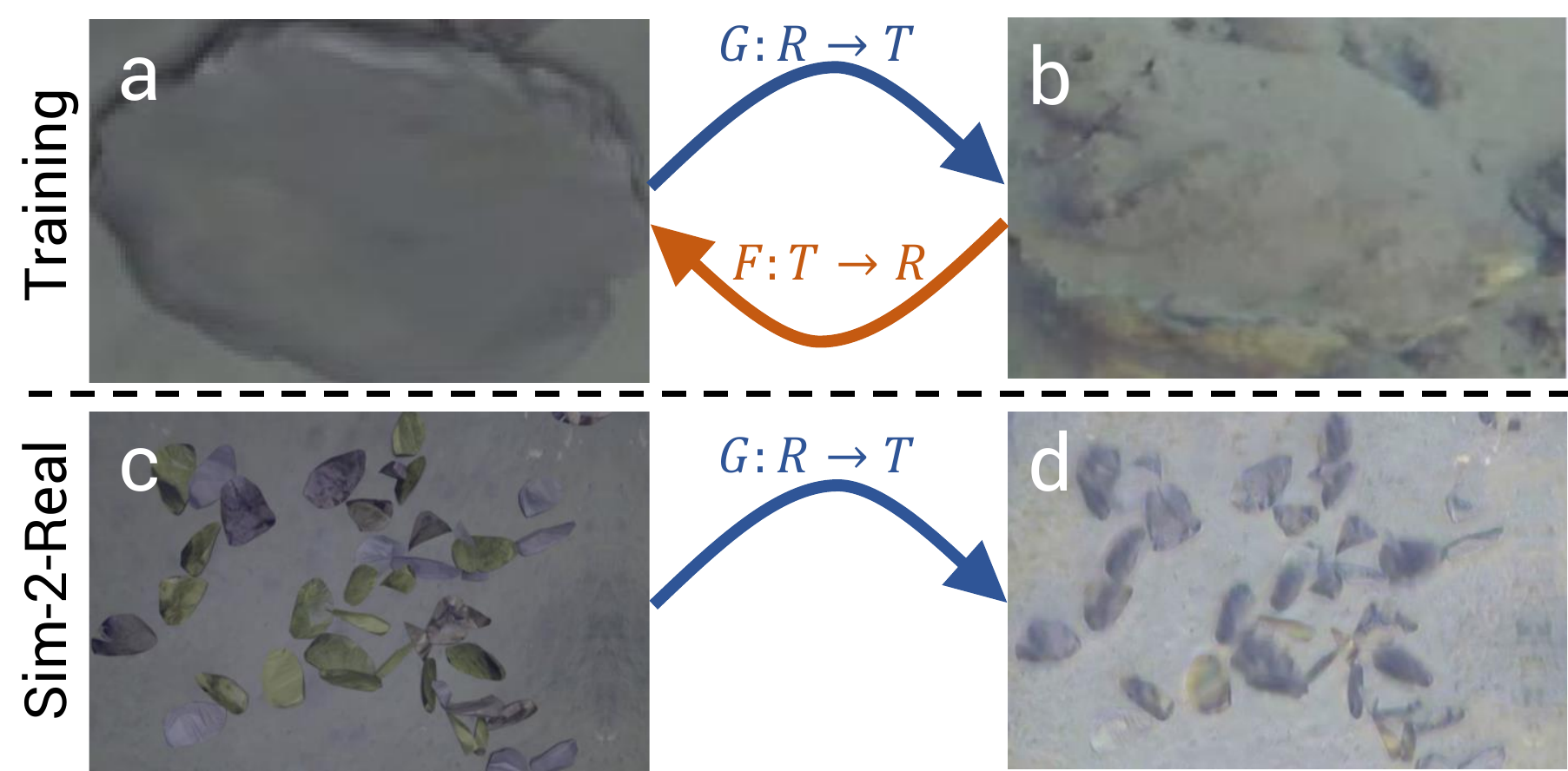}
\centering
\caption{Domain Adaptation to perform sim-to-real transfer. (a) A single sample of the simulated oyster, (b) A single sample of the real oyster, (c) Simulated oyster farm, (d) Synthetic oyster farm domain adapted to real world for photorealism.}
\label{fig:GAN_sample}
\end{figure} 
To render a realistic oyster image, we employ contrastive unpaired translation (CUT)~\cite{park2020contrastive} for unpaired image-to-image translation by learning functions to map from the Blender$^{\text{TM}}$~\cite{blender} domain $R$
to the target domain $T$. The overall loss function consists of three parts: GAN loss, PatchNCE loss, and ExternalNCE loss which are described next.\\[-5pt]

\textbf{GAN Loss}: Generators $G$ and $F$ are used to transfer
domains: $G : R \rightarrow T$ and $F : T \rightarrow R$. In order to improve image-to-image translation in CUT, ensuring the reconstructed
images $F(G(R)) \approx R$ is necessary. Thus, we want to  minimize the adversarial loss. Adversarial loss is defined by the following two components: discriminator loss and generator  loss. Where discriminator loss is to minimize the loss from misclassification between real and fake samples. As for the generator loss, the goal is to maximize the discriminator’s probabilities of being real. \\[-5pt]

\textbf{PatchNCE loss}: We first break the function $G$ into two parts, one encoder, and one decoder. $\hat{T} = G(R) = G_{dec}(G_{enc}(R))$.
$G_{enc}$ is used for image translation. Therefore, a patch of the input images can be represented as the feature stack of each layer, and the spatial location of the feature stack from the encoder $G_{enc}$. We want to ensure the cross-entropy between these feature stacks from different layers and spatial locations is minimized. Then the PatchNCE loss is introduced where NCE represents Noise Contrastive Estimation.\\[-5pt]

\textbf{ExternalNCE loss}:
Not only do we want to minimize PatchNCE loss, but we can also use the image patches from the rest of the dataset while training. A random negative image from the dataset is encoded  and is used to define externalNCE loss.\\[-5pt]

We refer the readers to \cite{park2020contrastive} for a detailed description of the Loss function and network. In our proposed system, the target domain is the realistic environment $T_{real}$.\\[-5pt]

\subsubsection{Details of Training the CUT}
We want the CUT network to be able to capture the synthetic oyster to real oyster translation. So, we extract 8959 images of single oysters from the synthetic data that we generated as $R$ and we also extract 957 samples of single oysters from the real underwater images as $T_{real}$. In the training phase, we learn two mapping functions: $G : R \rightarrow T$ and $F : T \rightarrow R$. Once the CUT is learned, we perform inference on our synthetic images to make them look photorealistic (See Fig. \ref{fig:GAN_sample}) by performing domain transfer which is further used to train our oyster detection network \textit{OysterNet}. In this work, we train CUT for 153 epochs to obtain our generator $G$.



\section{Experiments And Results}
\label{section:Experiments_and_results}
First, we describe the datasets used in this section. To validate our generated oyster model, we then compare the results obtained by two different segmentation networks and two different datasets. Lastly, we experiment with how different values of the oyster model parameters affect the segmentation results.

\subsection{Description of Datasets}
 \textbf{Rendered/Synthetic Dataset:} contains images obtained by randomly rendering generated 3D oyster models on the seabed using Blender$^{\text{TM}}$. Rather than just laying the oysters on the seabed, we simulate the randomization pose of the oysters in various realistic positions. This randomness in oyster pose makes the neural network more robust for recognizing the oyster in different poses. After the images are generated from Blender$^{\text{TM}}$, we used CUT, as described in Sec. \ref{section:problem_formulation}\textcolor{red}{-B}, to create
photo-realistic images. The rendered dataset contains 4800 images.\\[-5pt] 

\begin{figure}[t!]
\includegraphics[width=\linewidth]{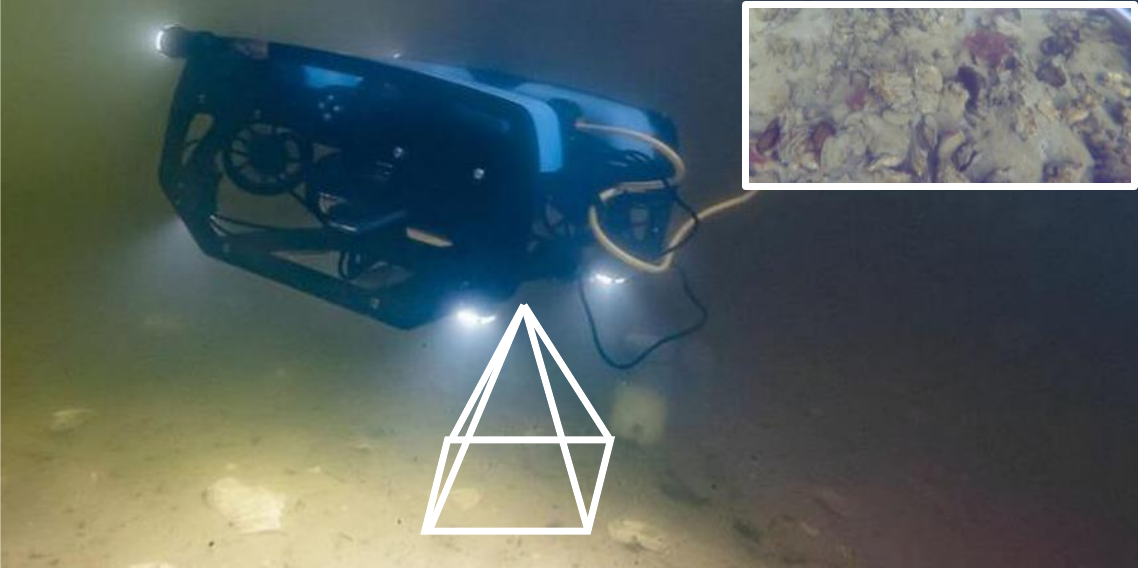}
\centering
\caption{BlueROV collecting data for our real dataset. The inset shows a sample image captured which is a part of our real dataset.}
\label{fig:oysteranddrone}
\end{figure}
 \textbf{Real Dataset:} contains images from Harris Creek taken by Chesapeake Bay Foundation by driving a BlueROV as shown in Fig. ~\ref{fig:oysteranddrone}. We labeled the images by hand. We have 29925 images of size $256\times 256$. There are over 3900 oysters labeled in the dataset. Unlike some other datasets for underwater object detection, oysters are really hard to recognize, even for humans. So, we worked with experts at the Chesapeake Bay Foundation to obtain this dataset.

\subsection{Experimental Results}
\textit{OysterNet} adapts UNet \cite{ronneberger2015u} as its backbone. We trained \textit{OysterNet} with a learning rate of 0.001 with decay. We use Adam optimizer coupled to the Jaccard loss \cite{bertels2019optimizing} loss function. We used a batch size is 32 and trained the network for 100 epochs.

We want to verify our hypothesis that the synthetic oyster data we generated can help improve the oyster semantic segmentation.

\cite{sadrfaridpour2021detecting} obtained the best results for oyster detection using the FPN\cite{lin2017feature} network. This is the model we use which we denote as the DCO (Detecting and Counting Oysters) method. In our observation  UNet~\cite{ronneberger2015u} architecture performed better using our approach which refer to it as \textit{OysterNet} (``Ours'') in Table \ref{tab:segmentation_result}. 

We want to use the minimal amount of real data for training as they are hard to obtain, expensive, and labor intensive. To this end, we use only 25\% of the real dataset we used as training which we will denote as $O_{real}$. We test our method on the remaining 75\% of the real dataset that we call $O$ in the Test Data. We denote all the images in our generated  synthetic dataset as $O_{syn}$ which will be used for training with/without real data. Formally,  $O_{syn\_and\_real}$ is the combination of $O_{real}$ and $O_{syn}$. 
\invis{We also perform a 25 and 75 split for Sadrfaridpour's Dataset~\cite{sadrfaridpour2021detecting} (labeled S in Tab. \ref{tab:segmentation_result}) for the training data and testing data. \\[-5pt]}

In this experiment, we perform training on $O_{real}$ (our real dataset) and test it on the $O$ (our held-out real dataset) with both \textit{OysterNet} and DCO methods. The Intersection over Union (IoU) scores are 18.16\% and 18.88\% respectively which serves as the baseline for oyster segmentation results for our dataset. Both the methods perform similarly in this case.
Then we want to observe the results when trained only on the generated synthetic dataset ($O_{syn}$). We train on $O_{syn}$ and test on $O$ with both methods. The IoU score is 7.45\% and 6.47\% respectively which is lower than our baseline. The network has learned to recognize oysters in the synthetic domain but the sim-to-real domain transfer is still not as desired. To this end, we add a small amount of real data to the synthetic data and use it for training ($O_{syn\_and\_real}$). We achieve a state-of-the-art IoU Score of \textbf{24.54\%} against the expert human labeled ground truth which is 35.1\% better than just using real dataset for training and 12.7\% better than DCO when trained on synthetic augmented real data.


\begin{table}[t!]
\caption{Comparison of Semantic Segmentation Results with the State-of-the-art.}
\centering
\begin{tabular}{llll}
\toprule
Method & Train Data & Test Data & IoU Score(\%)\\
\hline
DCO \cite{sadrfaridpour2021detecting}  & $O_{syn}$ & O & 6.47\\
DCO \cite{sadrfaridpour2021detecting} 	  &  $O_{real}$ & O & 18.88\\
DCO \cite{sadrfaridpour2021detecting}  & $O_{syn\_and\_real}$ & O & 21.76\\
\textit{OysterNet} (Ours) & $O_{syn}$ & O & 7.45\\
\textit{OysterNet} (Ours)  &  $O_{real}$ & O & 18.16\\
\textit{OysterNet} (Ours) & $O_{syn\_and\_real}$ & O & \textbf{24.54} \\

\bottomrule
\end{tabular}
\label{tab:segmentation_result}
\end{table}

\begin{table*}[t!]
\centering
\caption{Ablation Studies Of The Proposed Oyster Geometric Model (Also see Fig. \ref{fig:ablation_study}).}
\begin{tabular}{ p{1.5cm}|p{0.8cm}p{0.8cm}p{0.8cm}p{0.8cm}p{0.8cm}|p{1.5cm}|p{0.8cm}p{0.8cm}p{0.8cm}p{0.8cm}p{0.8cm}}
\toprule
\multicolumn{6}{c|}{\textbf{Scenario a}} & \multicolumn{6}{c}{\textbf{Scenario b}} \\
\hline
\textbf{$\mu_1$} &50 &100 &150 &200 &250 & \textbf{$\mu_2$}&50 &100 &150 &200 &250\\
\hline
\textbf{IoU(\%) }&  21.70 & 23.10 & 24.54  & 23.21 & 17.67 &\textbf{IoU(\%)} &24.19 & 21.23 & 24.54 & 22.34 & 19.57\\
\hline
\multicolumn{6}{c|}{\textbf{Scenario c}} & \multicolumn{6}{c}{\textbf{Scenario d}} \\ 
\hline
\textbf{$\sigma_1$} &50 &100 &150 &200 &250 & \textbf{$\sigma_2$} &15 &30 &60 &120 &240\\
\hline
\textbf{IoU(\%)} &  19.37 & 22.07 & 24.54  & 23.64 & 20.67  &\textbf{IoU(\%)} &24.54 &21.53 & 23.64 & 22.42 & 22.95 \\
\hline
\multicolumn{6}{c|}{\textbf{Scenario e}} & \multicolumn{6}{c}{\textbf{Scenario f}} \\
\hline
\textbf{$\alpha$} &  5-10 & 10-15 & 15-20  & 20-25 & 25-30 & \textbf{Scale(\%) }& 10-15 & 15-20  & 20-25 & 25-30 &30-35 \\
\hline
\textbf{IoU(\%)} &  22.76 & 22.88 & 24.54  & 22.63 & 22.66  &\textbf{IoU(\%)} & 18.97 & 23.21  & 24.36 & 24.54 &21.20 \\
\bottomrule
\end{tabular}
\label{tab:AblationStudy}
\end{table*}


\invis{\textbf{Experiment 2:} In order to show the difference between the two datasets, we perform the cross-training on datasets $O$ and $S$. Training on $S$ and testing on $O$ We see the IoU Score near 3\%, which already indicates that  datasets $O$ and $S$ are very different from one another. 
Further, we performed the training on $O_{real}$  and test it on $S$. The IoU Score dropped significantly to nearly 20\% compared to 55.02\% with training on $S$ only.  
Even though both datasets are for oyster segmentation/detection, the datasets are very different from each other. There are different types of oysters and different living environments which cannot be presented by datasets from other locations. However, we proposed an approach that can systematically model the oyster and generated datasets for oysters. }

 The results are tabulated in Table \ref{tab:segmentation_result}. It shows that with the added synthetic dataset, we achieve a significant $\sim$35.1\% accuracy improvement over just using the real dataset for training. As we can see from Figs.~\ref{fig:detection_result}\textcolor{red}{b} and \ref{fig:detection_result}\textcolor{red}{e}, the network predicts mostly sediments as oysters when trained using only real data. This leads to a lot of false positives and false negatives. However, when the network is trained  using real data augmented with our synthetic data (which we call \textit{OysterNet}), there is a significant decrease in false predictions as we can see from Figs.~\ref{fig:detection_result}\textcolor{red}{c} and \ref{fig:detection_result}\textcolor{red}{f}.



Since the generated dataset affects accuracy significantly, we ablate on how different parameters in our proposed geometric model affect performance in the next section.

\subsection{Ablation Studies}
We varied the parameters $\Theta$ that controlled amount of high-frequency components and the number of layers to generate oyster models. We also varied the size of the oyster when rendering it in the simulation. When not specified, the parameters are set as $\mu_1=150$, $\mu_2=150$, $\sigma_1=150$, $\sigma_2=15$, $\alpha \in [15,20]$, and $ Scale \in [25,30]$. Only one parameter is varied at a time while keeping others fixed to the values specified. No two oysters are the same with respect to size and height, therefore we chose different ranges for $\alpha$ and Scale. 
We see from Table \ref{tab:AblationStudy}\textcolor{red}{a} that IoU Score increases with mean of the noise and then decreases. 
\vspace{5mm}
\begin{figure*}[ht!]
\includegraphics[width=\textwidth]{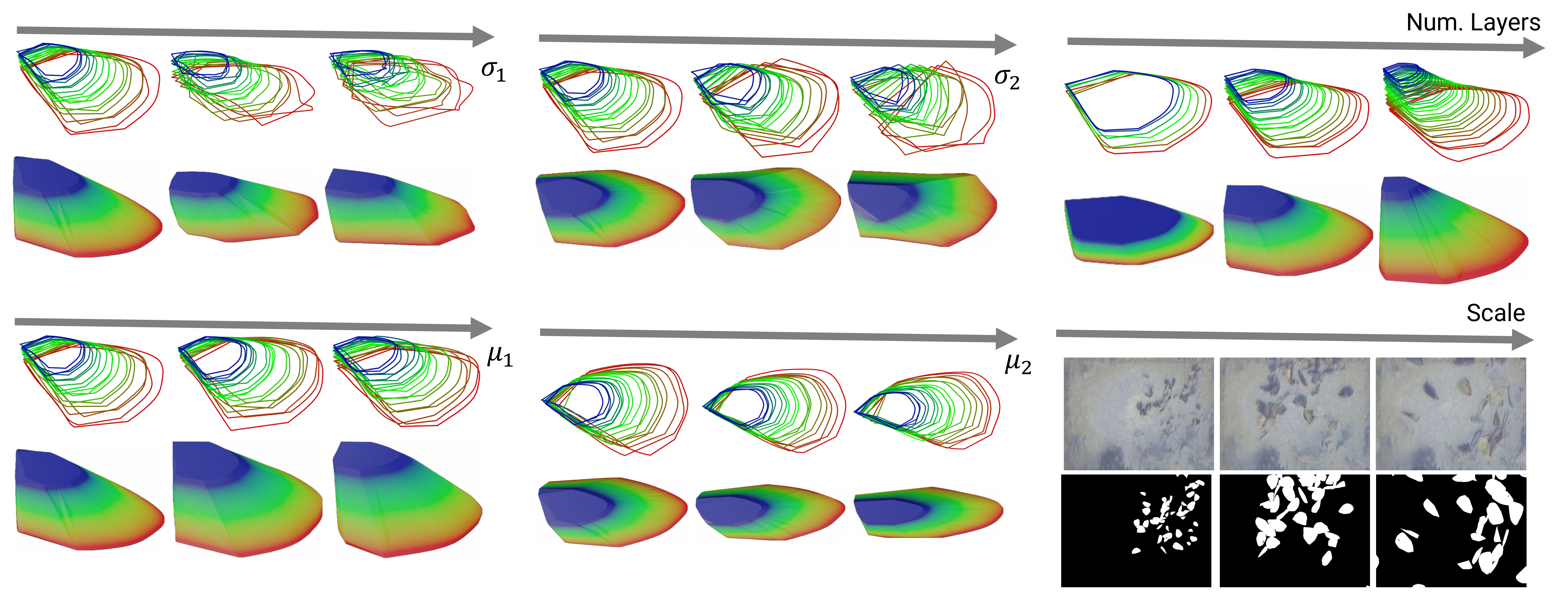}
\centering
\caption{Qualitative demonstration of how different parameters of our oyster model affect the shape of the oyster. The parameters are in the same order as the experiments in Table \ref{tab:AblationStudy}.}
\label{fig:ablation_study}
\end{figure*}

This is because the noise increases, and the variety of the oysters increases which would benefit the detection of the oysters. However, when the noise is too large ($\mu_1 > 150$), the oyster generated no longer looks like an oyster (See Fig. \ref{fig:ablation_study}). The IoU Score drops significantly when $\mu_1 = 250$. A similar trend is observed with $\mu_2$ and $\sigma_1$ with the IoU Score peaking around 150 (Table \ref{tab:AblationStudy}\textcolor{red}{b} and \ref{tab:AblationStudy}\textcolor{red}{c} ). We also observe that with
increase in standard deviation ($\sigma_2$) for $X_2$ (Table \ref{tab:AblationStudy}\textcolor{red}{d}), the detection accuracy drop slightly. By varying the number of layers ($\alpha$) for the oysters, we notice that the change in the IoU Score is relatively minimal. Since the image that we generated is in 2D, the parameter for layers ($\alpha$) does not have a significant effect (Table \ref{tab:AblationStudy}\textcolor{red}{e}) as we mostly observe oysters from far away and in mostly top view.

In Table \ref{tab:AblationStudy}\textcolor{red}{f}, we observe that the scale of the oyster has to match the size of the oyster in the image to get the optimal result. The accuracy drops when the scale of the oyster is either too large or too small. The IoU Score difference between the scale values in the ranges of $20-25$ and $25-30$ is only $0.18\%$ (Tables~\ref{tab:AblationStudy}\textcolor{red}{f}).

\section{CONCLUSIONS}
\label{section:Conclusions}
In this work, we first model the geometry of oyster shells and render the oyster images in a game engine. Then we perform an image-to-image transformation from the simulation domain to the real-world domain. With the help of the generated synthetic dataset, when augmented to the real dataset we showed an improvement in the semantic segmentation IoU score for the oysters by 35.1\% over just using real data for training and 12.7\% over the current state-of-the-art. These results highlight that for data-critical applications when collecting real images is challenging, it is possible to model the images using the underlying geometry of the object to create photorealistic images that will improve object detection drastically. To the best of our knowledge, this is the first attempt to model 3D structure of oysters. Being the first work in this field, there are many directions and possible improvements that can be made to our \textit{OysterNet} framework to increase the accuracy of semantic segmentation. One possible improvement will be to model both shells of the oysters instead of a single shell. As for the detection phase, new network architectures can be explored to tackle this problem. Finally, more models of the oysters along with additional shape noise could be utilized in the future for a more robust detection.




\addtolength{\textheight}{-10cm}   


\section*{ACKNOWLEDGMENT}
We would like to thank to Patrick Beall and Doug Myers from Chesapeake Bay Foundation for annotation some of the images for us. We would also like to thank Allen Pattillo and Chahat Deep Singh from University of Maryland for proofreading our manuscript. 

\bibliographystyle{IEEEtran}
\bibliography{IEEEabrv,refs}

\end{document}